\documentclass[letterpaper, 10 pt, conference]{ieeeconf}  

\IEEEoverridecommandlockouts                              

\usepackage{graphics} 
\usepackage{epsfig} 
\usepackage{mathptmx} 
\usepackage{times} 
\usepackage{amsmath} 
\usepackage{amssymb}  
\usepackage{cite}
\usepackage{color}
\usepackage{amsfonts} 

\usepackage{times}
\usepackage{soul}
\usepackage{url}
\usepackage{graphicx}
\usepackage{amsmath}
\usepackage{amssymb}
\usepackage{booktabs}
\usepackage{multirow}
\usepackage{threeparttable}
\usepackage{makecell}
\usepackage{color}
\usepackage{verbatim}

\title{\LARGE \bf
MonoPGC: Monocular 3D Object Detection with \\Pixel Geometry Contexts
}

\author{Zizhang Wu$^{1}$, Yuanzhu Gan$^{1}$, Lei Wang$^{1}$, Guilian Chen$^{1}$, Jian Pu$^{2}$
\thanks{$^{1}$Zongmu Technology}
\thanks{$^{2}$Fudan University}
}

\begin{document}

\maketitle
\thispagestyle{empty}
\pagestyle{empty}


\begin{abstract}

Monocular 3D object detection reveals an economical but challenging task in autonomous driving.
Recently center-based monocular methods have developed rapidly with a great trade-off between speed and accuracy, where they usually depend on the object center’s depth estimation via 2D features. 
However, the visual semantic features without sufficient pixel geometry information, may affect the performance of clues for spatial 3D detection tasks.
To alleviate this, we propose \textbf{MonoPGC}, a novel end-to-end \textbf{Mono}cular 3D object detection framework with rich \textbf{P}ixel \textbf{G}eometry \textbf{C}ontexts.
We introduce the pixel depth estimation as our auxiliary task and design depth cross-attention pyramid module (\textbf{DCPM}) to inject local and global depth geometry knowledge into visual features.
In addition, we present the depth-space-aware transformer (\textbf{DSAT}) to integrate 3D space position and depth-aware features efficiently.
Besides, we design a novel depth-gradient positional encoding (\textbf{DGPE}) to bring more distinct pixel geometry contexts into the transformer for better object detection.
Extensive experiments demonstrate that our method achieves the state-of-the-art performance on the KITTI dataset.

\end{abstract}

\section{INTRODUCTION}

Accurate detection of 3D objects in various scenarios has extensive applications such as autonomous driving and robotic manipulation~\cite{reading2021categorical,liu2020smoke}. %
To obtain a precise perception of 3D information~\cite{guo2021liga}, many 3D object detection methods
rely on 3D point clouds or stereo images that typically require costly system setups, such as LiDAR or stereo sensors\cite{li2021monocular,brazil2019m3d}.
In contrast, 
monocular 3D object detection, which only uses a monocular RGB camera as a simpler and cheaper setting for deployment, has attracted increasing attention recently.

Most existing monocular 3D object detection methods can be roughly divided into two categories.
The pseudo-LiDAR-based methods~\cite{wang2019pseudo, reading2021categorical} normally lift learned features from 2D space to 3D space and then conduct 3D object detection. 
The center-based methods~\cite{zhou2019objects, li2020rtm3d, li2021monocular, wang2021probabilistic, yin2021center} directly adopt image features to predict object's center location, orientation, dimension, depth and other auxiliary monocular tasks.
Generally, the pseudo-LiDAR-based methods perform more accurately yet time-consuming for lifting images to 3D space. In contrast, the center-based methods achieve better speed-accuracy trade-off \cite{li2021monocular, liu2021learning}.
However, the performance of center-based methods heavily depends on the estimation of depth. 
When incorrect estimation occurs, localization error easily affects the performance of the following detection task \cite{guo2021liga}.
As a consequence, without considering 3D geometry information, the center-based methods are often inferior to those pseudo-LiDAR-based methods \cite{wang2019pseudo, weng2019monocular, reading2021categorical}.
While MonoDTR \cite{huang2022monodtr} introduces the pixel-level depth-aware auxiliary supervision to achieve great improvement, its pixel geometry information still appears some indistinct and smooth for object detection task, which may bring disturbance with different semantics of the same depth.

To address the issues of coarsely encoded 3D pixel geometry information, we propose \textbf{MonoPGC}, a novel end-to-end monocular 3D object detection with rich pixel geometry contexts, and surpass the prior state-of-the-art 3D monocular detection methods.
Specifically, we introduce the pixel-level depth estimation as an auxiliary task~\cite{huang2022monodtr}, and design the Depth Cross-attention Pyramid Module (DCPM) to obtain more precise depth prediction from the multi-scale image features. 
Inspired by implicit neural representation (INR) methods \cite{mildenhall2020nerf, liu2022petr, zhou2022cross},
we further propose to encode 3D coordinates information into depth-aware features by the Depth-Space-Aware Transformer (DSAT), for enhancing the 3D position-aware pixel geometry information. 
In addition, we introduce a new term, depth gradient, for encoding the object boundary and
propose the Depth-Gradient Positional Encoding (DGPE) to spotlight edges of objects, which serves as valuable hints for pixel geometry modeling. 
We summarize our contributions as follows:

(1) We propose a novel framework, MonoPGC, leveraging  pixel-level geometry information to boost the 3D object detection. 
To achieve so, we introduce the pixel depth estimation to strengthen geometry knowledge. Moreover, we design the cross-attention pyramid module to avoid incorrect depth priors, and produce more robust depth-aware features integrated with local and global receptive field. 

(2) We present the first depth-space-aware transformer to integrate 3D space coordinates and depth aware features efficiently. 
We propose a novel depth-gradient positional encoding (DGPE) to bring more distinct and rich hints into the transformer for the object detection task.

(3) Our MonoPGC framework achieves the state-of-the-art performance on the KITTI \cite{kitti} dataset. 
Particularly, extensive experiment results illustrate that it achieves 24.68$\%$, 17.17$\%$, and 14.14$\%$ in terms of the AP$_{40}$ metric, on the easy, moderate, and hard test setting for Car category, respectively.

\section{RELATED WORK}

\begin{figure*}[t]
    \centering
    \includegraphics[width=0.8\textwidth]{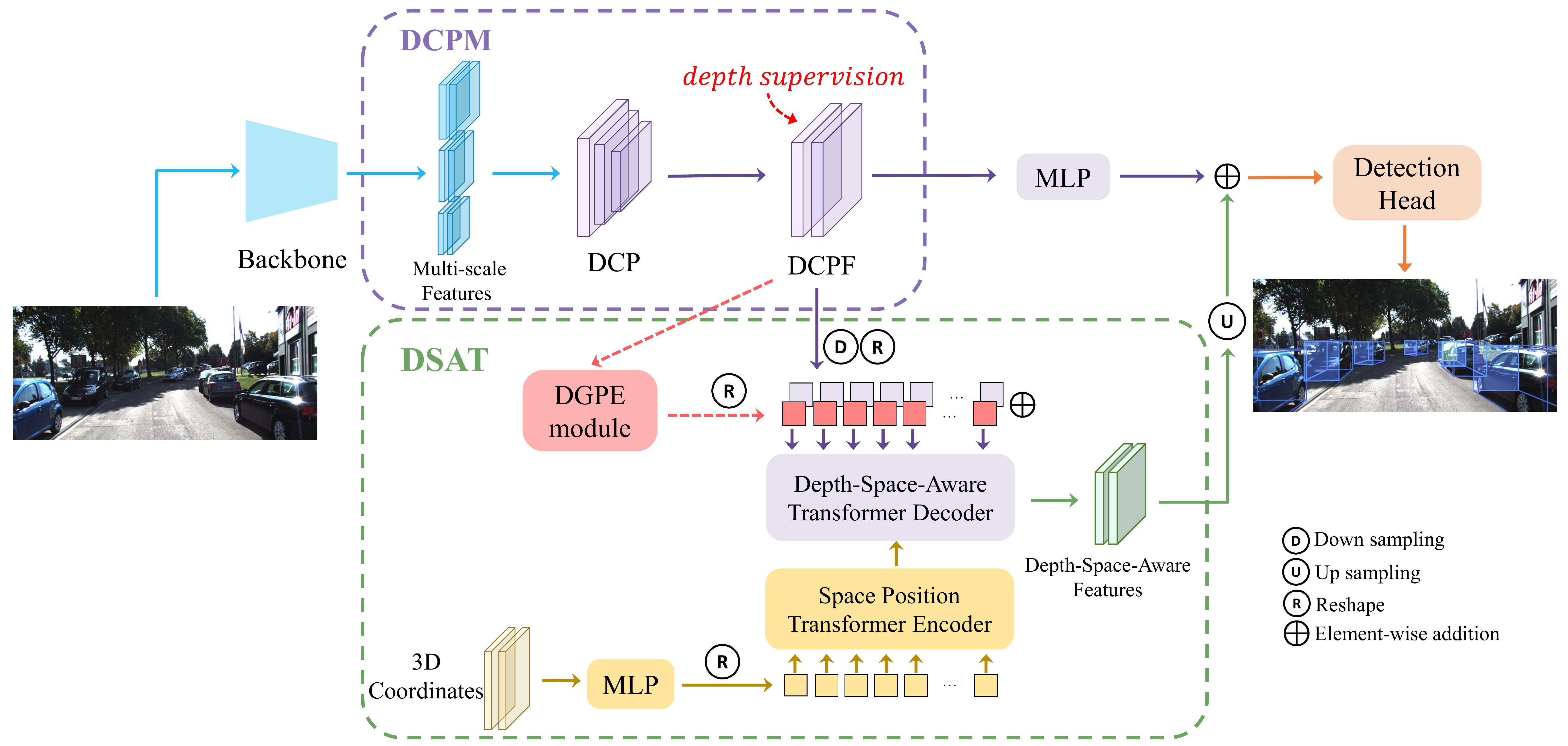}\\
    \vspace{-0.1in}
     \caption{The framework of our MonoPGC, which consists of the backbone, the depth cross-attention pyramid module (DCPM), the depth-space-aware transformer (DSAT) module, and the detection head.
     ``DCP'' denotes the depth cross-attention pyramid, ``DCPF'' denotes depth cross-attention pyramid features, and ``DGPE'' denotes the  depth-gradient positional encoding.
     }
    \label{framework1}
    \vspace{-0.2in}
\end{figure*}

\subsection{Monocular 3D Object Detection}
Many prior works \cite{chen2016monocular}\cite{kehl2017ssd}\cite{ku2019monocular}\cite{brazil2019m3d}\cite{simonelli2019disentangling}\cite{liu2021ground} have explored the inherently ill-posed problem of detecting 3D objects from monocular images.
Due to the lack of depth information from images, monocular 3D detection significantly falls behind Lidar-based and stereo-based counterparts.
Many works \cite{liu2019deep}\cite{cai2020monocular}\cite{zhang2021objects} ease this problem by utilizing 2D-3D geometric constraints to improve 3D detection performance.
Deep3Dbox \cite{mousavian20173d} proposes MultiBin loss and enforces constraint between 2D and 3D boxes with geometric prior.
Mono3D \cite{chen2016monocular} is a seminal work that uses region proposals to deal with semantics and contexts.
CenterNet \cite{zhou2019objects} proposes a center-based anchor-free method but with restrained accuracy.
Following this work, center-based series SMOKE \cite{liu2020smoke}, KM3D \cite{li2021monocular} and RTM3D \cite{li2020rtm3d} assist the regression of object depth by solving a Perspective-n-Point method and have achieved remarkable results.
However, existing works, computing 3D locations explicitly based on 2D predictions, usually suffer from the well-known error amplification effect.
Recent works \cite{lu2021geometry,zhang2021objects,ma2021delving,liu2021learning} try to use uncertainty modeling \cite{lu2021geometry}, sophisticated model ensemble \cite{zhang2021objects} or auxiliary monocular contexts \cite{liu2021learning} to improve the performance.

\subsection{Transformer-based Object Detection}
Many recent works \cite{wang2022detr3d}\cite{chen2022graph}\cite{liu2022petr}\cite{li2022bevformer}\cite{huang2022monodtr} take a step further and explore the usage of transformer in 3D object detection.
DETR \cite{carion2020end} brings Transformer into the field of object detection.
DETR3D \cite{wang2022detr3d} proposes a new paradigm to address the ill-posed inverse problem of recovering 3D information from 2D images.
Graph-DETR3D \cite{chen2022graph} improves the DETR3D by a novel dynamic graph feature aggregation module.
BEVFormer \cite{li2022bevformer} learns a unified BEV representation with spatiotemporal transformers.
MonoDTR \cite{huang2022monodtr} proposes to globally integrate context- and depth-aware features with transformers and inject depth hints into the transformer.

\subsection{Implicit Neural Representation}
Early work has investigated the implicit neural representation of continuous 3D shapes by employing deep networks that map coordinates to signed distance functions \cite{park2019deepsdf}\cite{jiang2020local} or occupancy fields \cite{genova2020local}\cite{mescheder2019occupancy}.
NeRF \cite{mildenhall2020nerf} utilizes multi-layer perceptrons (MLPs) to encode 5D radiance fields (3D volumes with 2D view-dependent appearance), which is referred to as positional encoding and proved to be an efficient way for modeling complex 3D scenes.
Inspired by the success of NeRF\cite{mildenhall2020nerf}, recent works \cite{liu2022petr}\cite{zhou2022cross} take similar approach mapping both 3D coordinates and camera parameters (camera intrinsics and extrinsics) to positional encoding.
These methods can be regarded as an extension of INR in 3D object detection \cite{liu2022petr}.
Furthermore, they also use the popular Transformer architecture \cite{vaswani2017attention} to exploit the encoded information.

\begin{figure*}[t]
    \centering
    \includegraphics[width=0.7\textwidth]{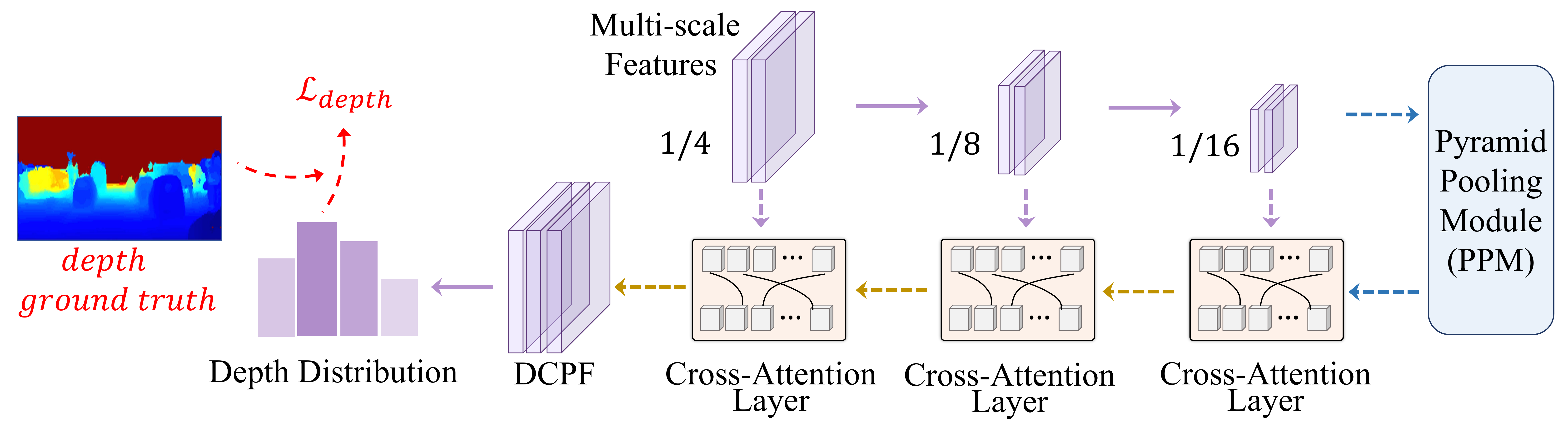}\\
    \vspace{-0.1in}
     \caption{The framework of Depth Cross-attention Pyramid Module (DCPM).
     We fuse the multi-scale features and introduce depth supervision to obtain the depth loss.}
    \label{dpf}
    \vspace{-0.2in}
\end{figure*}

\section{Method}
\vspace{-0.1in}
\subsection{The Overview}
Figure \ref{framework1} demonstrates the framework of our MonoPGC, which mainly consists of four components: the backbone, the depth cross-attention pyramid module (DCPM), the depth-space-aware transformer (DSAT) module, and the detection head.
Given an image $I \in R^{3\times H_I\times W_I}$, we adopt the backbone network (DLA34 \cite{yu2018deep}) to extract its multi-scale 2D features $F^{2d}_i=\left \{{F_i^{2d}\in R^{C\times H_{F_{i}}\times W_{F_{i}}}, i=\frac{1}{4}, \frac{1}{8},\frac{1}{16}} \right \}$. 
The depth cross-attention pyramid module (DCPM) fuses multi-scale features for pixel depth estimation with the contribution of cross-attention and pyramid designs, then achieves DCP features (DCPF) (Section \ref{section3.2}).
Furthermore, the 3D world space coordinates from the 3D coordinates generator, could be a strong indicator for the implicit neural representation.
Thus, we integrate DCPF and space coordinates by the depth-space-aware transformer (DSAT) module and insert the informative depth-gradient positional encoding into the transformer through the depth-gradient positional encoding (DGPE) module (Section \ref{section3.3}).
Finally, we apply DSA features to the detection head for object detection task (Section \ref{section3.4}). 

\subsection{Depth Cross-attention Pyramid  Module}\label{section3.2}
Existing monocular methods \cite{liu2020smoke,wang2021probabilistic,zhang2021objects,liu2021learning} usually estimate the depth of objects' center, which may bring inaccurate depth priors when objects stand in various poses and headings.
Pixel-wise depth shall be helpful, since accurate pixel object depth bring extra geometry context, such as it's less possible to appear objects in a large depth.

In our work, we regard the pixel depth estimation as our auxiliary task, injecting the depth information to the visual features.
We propose our depth cross-attention pyramid module (DCPM). 
Pyramid structure combined with cross-attention layer on different scales, improves the receptive field and multi-scale perception for better depth estimation. 

\noindent \textbf{Depth Cross-attention Pyramid Feature.}
Figure~\ref{dpf} briefly shows the architecture of our depth cross-attention pyramid module. 
The input of this module remains the multi-scale features from the backbone.
Firstly, we adopt the pyramid pooling module (PPM) \cite{zhao2017pyramid} to aggregate the global and local information of the whole image from the top-level feature maps.
Sequentially, we adopt three cross-attention layers to fuse the multi-level features, which finally outputs the enhanced depth cross-attention pyramid features $F_{DCP} \in R^{D\times H_{\hat{F}}\times W_{\hat{F}}}$, where $H_{\hat{F}}$ and $W_{\hat{F}}$ denote 0.25 scale of input image's size.
\begin{equation}
F_{DCP}=DCPM(F^{2d}_{i}), i=\frac{1}{4}, \frac{1}{8},\frac{1}{16}
\end{equation}
By utilizing the cross-attention layers, the module can extract non-local context features with global and local perception and realize the efficient multi-level feature fusion.
We intercept the cross-attention block from Linear transformer \cite{katharopoulos2020transformers} as our cross-attention layer.
We leave the detailed architecture of cross-attention layers in the supplementary material.

\begin{figure}[t]
    \centering
    \includegraphics[width=0.45\textwidth]{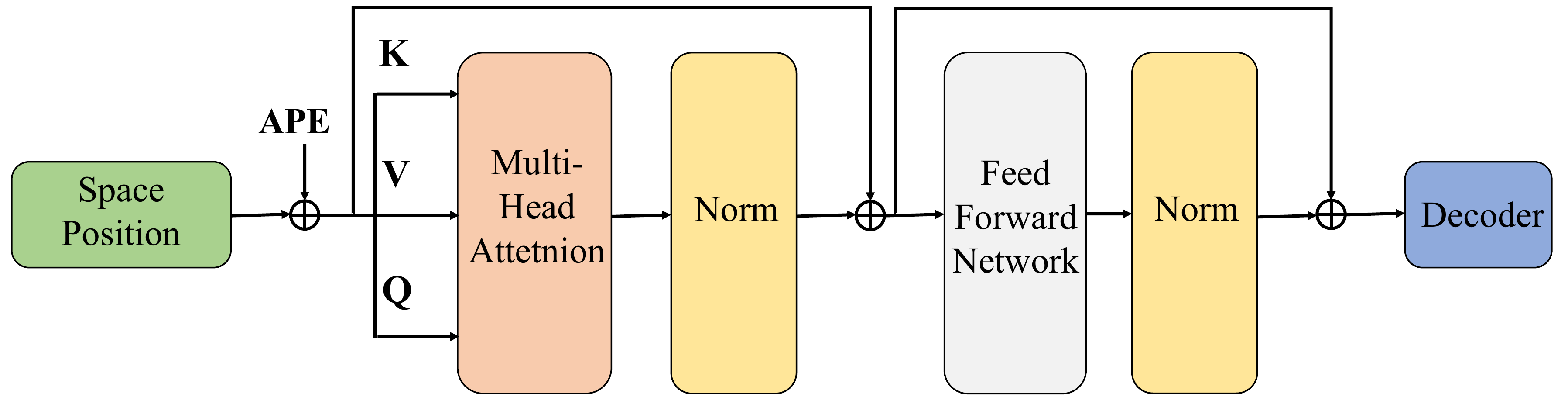}\\
     \caption{The encoder of Depth-Space-Aware Transformer (DSAT) with space position as input. We adopt absolute positional encoding (APE) \cite{VIT} as encoder's positional encoding.}
    \label{transformer_encoder}
    \vspace{-0.2in}
\end{figure}

\noindent \textbf{Depth estimation.}
As illustrated in Figure~\ref{framework1}, accepting the depth cross-attention pyramid features (DCPF), we adopt a simple 3$\times$3 convolution layers to achieve depth estimation.
Following \cite{reading2021categorical,huang2022monodtr}, we predict the depth distribution: the probability of discretized depth
bins $D_{gt} \in R^{D\times H_{\hat{F}}\times W_{\hat{F}}}$, where D is the number of depth bins.
We adopt linear-increasing discretization (LID) \cite{tang2020center3d}, discretizing the depth ground truth to the depth bins. The relation between bin index and depth value states as follows: 
\begin{equation}\label{equation_LID}
    d_c = d_{min} + \frac{d_{max} - d_{min}}{D(D+1)}\cdot d_i(d_i+1)
\end{equation}
where $d_i$ denotes the bin index and $d_c$ denotes the correspond depth value.
Then we exploit the focal loss for the auxiliary depth task, trying to classify the depth to a correct depth bin and pay more attention to the foreground object pixels \cite{reading2021categorical}.
\begin{equation}
    L_{depth} = \frac{1}{W_{\hat{F}} \cdot H_{\hat{F}}}\sum_{u=1}^{W_{\hat{F}}}\sum_{v=1}^{H_{\hat{F}}}FocalLoss(D_{pred}(u,v), D_{gt}(u,v))
\end{equation}
\subsection{Depth-Space-Aware Transformer}\label{section3.3}
Implicit neural representation (INR) proves to be an efficient way for 3D objects or 2D images modeling by mapping the coordinates to visual signal.
Transformer architecture reveals a powerful weapon for self-modeling or cross-modeling.
In this section, we exploit the effective transformer encoder-decoder structure to map the 3D space coordinates to our visual DCPF, enhancing the pixel's geometry contexts.

\begin{figure}[t]
    \centering
    \includegraphics[width=0.48\textwidth]{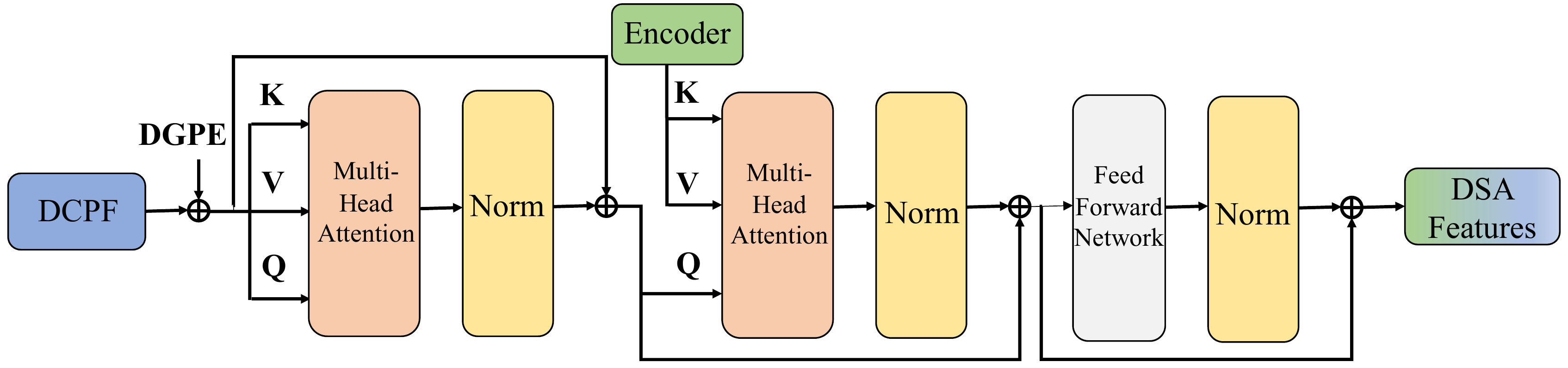}\\
     \caption{The decoder of Depth-Space-Aware Transformer (DSAT) with DCPF and encoder' output as input. Depth-gradient Positional Encoding (DGPE) assists pixel geometry modeling.}
    \label{transformer_decoder}
    \vspace{-0.2in}
\end{figure}

\begin{figure*}[t]
    \centering
    \includegraphics[width=0.65\textwidth]{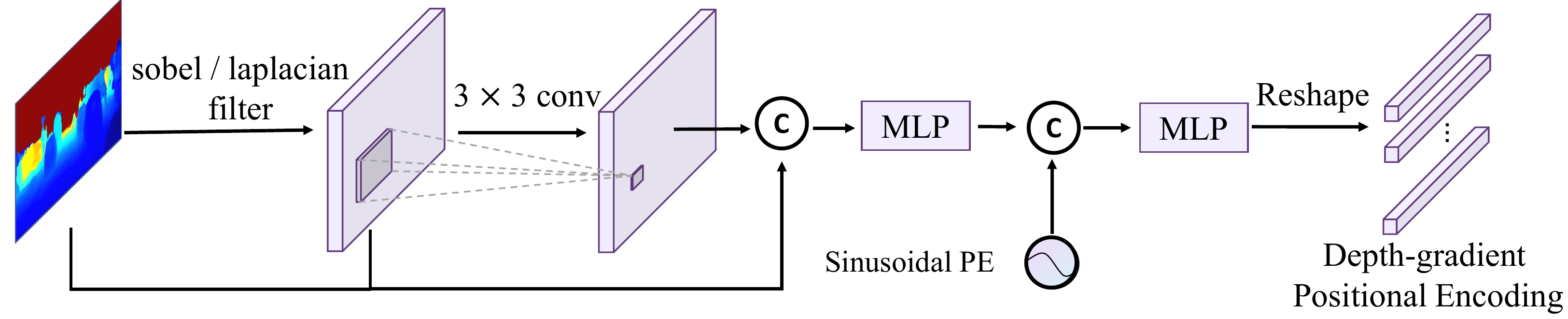}\\
     \caption{The structure of Depth-Gradient Positional Encoding (DGPE). 
     The input reveals our depth prediction in above DCPM. 
     The operation of C indicates concatenation. The channels after the 3x3 convolution are three, of two channels for the sobel filter in X and Y axes and one channel for laplacian filter.}
    \label{DCPE}
    \vspace{-0.2in}
\end{figure*}

\noindent \textbf{3D Coordinates Grid.}
Firstly, we need to obtain the 3D coordinates. Since coordinates are continuous, following DSGN \cite{chen2020dsgn}, we discrete the 3D coordinates to the integer grid and project the points from camera frustum space to 3D space. 
Specifically, every pixel coordinate in the image indicates $P_{j}^{i}=(u_{j}, v_{j}, 1, 1) ^{T}$, where $(u_{j} , v_{j})$ means the pixel coordinate in the image.
Following \cite{reading2021categorical, huang2022monodtr}, we also apply LID \cite{tang2020center3d} to transform the depth within ROI from continuous space to discretization intervals.
Then we recovery the discrete bin indexes to depth values $d_{1},d_{2},..  .,d_{N}$ (N denotes the number of depth bin), which acts as our depth grid coordinates along the axis orthogonal to the image plane. The camera frustum space points turn to $P_{j}^{m}=(u_{j} \times d_{j}, v_{j} \times d_{j},d_{j},1) ^{T}$, $d_{j}\in d_{1},d_{2},..  .,d_{N}$.  
Consequently, we reversely project our camera frustum space points to the 3D world space with the assistance of intrinsics and extrinsics calibration:
\begin{equation}
P_{j}^{3d}=K_{E}^{-1}K_{I}^{-1}P_{j}^{m}
\end{equation}
where $K_{E}\in R^{4\times 4}$ and $K_{I}\in R^{4\times 4}$ indicate the extrinsics and intrinsics calibration matrix, and $P_{j}^{3d}=(x_{j},y_{j},z_{j},1)^{T}$ denotes 3D coordinates grid calculated from camera frustum space coordinates grid.
After normalizing and reshaping, we achieve the normalized 3D coordinates grid $P^{3d}=\left \{{P_{j}^{3d}\in R^{(D\times4)\times H_{\hat{F}}\times W_{\hat{F}}}} \right \}$.

\noindent \textbf{Space Position Encoder.}
Implicit neural representation (INR) usually adopts MLPs as the mapping function \cite{mildenhall2020nerf}.
Inspired by rapid-developed transformer architecture, which succeeds in modeling the long-range relationships, we adopt the encoder-decoder structure to map the 3d space coordinates to our visual features.

As shown in Figure \ref{framework1}, firstly we deliver our 3D coordinate grid to the MLPs for simply linear transformation.
Sequently, we flatten them to $\hat{P}^{3d}\in R^{N\times (D\times4)}$, where $N = H_{\hat{F}} \times W_{\hat{F}}$ is the input of the space position (SP) transformer encoder.
\begin{equation}
    F_{E}=Encoder(\hat{P}^{3d})
\end{equation}
To mitigate the time and memory budget, following \cite{huang2022monodtr}, we also utilize the Linear
transformer \cite{katharopoulos2020transformers} to accelerate the attention operation. Figure \ref{transformer_encoder} shows our encoder framework, which consists of transformer self-attention modules.

\noindent \textbf{Depth-Space-Aware Decoder.}
We adopt the transformer decoder to complete the space coordinates mapping to our visual features DCPF.
Figure \ref{transformer_decoder} demonstrates our decoder structure.
We feed our DCPF into the decoder instead of learnable embeddings (object query).
Simple learnable embedding can't fully represent complex scale variant situations due to the perspective projection \cite{huang2022monodtr}.
Besides, our DCPF contains depth knowledge, which is more suitable for the transformer decoder to enhance the coordinate mapping.

The encoder's output embeds into the decoder's cross-attention modules and the multi-layer transformer decoder efficiently models the transformation between space coordinates and DCPF, resulting in a more robust mapping. The output of decoder denotes depth-space-aware features (DSAF).
\begin{equation}
    F_{DSA}=Decoder(F_{DCPF},F_{E})
\end{equation}

\noindent \textbf{Depth-Gradient Positional Encoding (DGPE).}
To increase position information for transformer's input, adding position embedding to features becomes conventional operation in transformer, such as sinusoidal positional encoding \cite{vaswani2017attention} and absolute position encoding \cite{VIT}. 
MonoDTR \cite{huang2022monodtr} regards depth as hints to produce depth positional embedding.
However the depth usually appears indistinct and smooth for object detection task, and is affected with same-depth-different-semantics confusion, which limits the depth's potential.

We propose a new hint, depth gradient, for encoding more clues for the object detection task.
Our depth-gradient positional encoding (DGPE) introduces the gradient information to spotlight the edge of objects, which act as valuable indications for anchor-free object detection.
Specifically, we adopt sobel and laplacian filters, which are common detectors for image edge detection.
Sobel filter calculates the first derivatives of the image separately for the X and Y axes and laplacian filter calculates second order derivatives in a single pass.
As shown in Figure \ref{DCPE}, we manipulate our depth prediction with two gradient-based edge filters.
We also adopt a 3$\times$3 convolution layer for local perception.
Finally we combine all the edge features, depth knowledge with the the init positional embedding, to produce our depth-gradient positional encoding (DGPE).
\subsection{Detection Head and Loss}\label{section3.4}
We reserve the detection head and loss from \cite{liu2021learning}.
Specifically, we adopt the anchor-offset style during training.
We predict the class-specific heatmap for the 2D bounding box's center, orientation, dimension, object depth, depth uncertainty and other auxiliary monocular tasks.
We also maintain the loss calculation, added with our depth loss.
\cite{liu2021learning} reveals more details about the 3D detection heads and losses.
\begin{equation}
    L=\lambda_{depth}L_{depth} + \lambda_{cls}L_{cls} + \lambda_{reg}L_{reg}
\end{equation}
\vspace{-0.3in}





\vspace{0.1in}
\section{Experiments}
\subsection{Dataset, metrics and implementation}
\noindent \textbf{KITTI.}
The popular KITTI \cite{kitti} dataset officially consists of 7,481 samples for training and 7,518 for testing.
It is a common practice \cite{reading2021categorical} to divide training samples into a training set with 3,712 samples and a validation set with 3,769 samples.
We conduct ablation studies on the validation split with models trained on the training split.
The 3D Average Precision ($AP_{3D}$) and BEV Average Precision ($AP_{BEV}$) are two vital evaluation metrics for KITTI 3D object detection.
Following previous methods \cite{reading2021categorical}, our approach is compared on the test set using AP$_{40}$ instead of AP$_{11}$. 
In addition, the Car category has an IoU threshold of 0.7 while the Pedestrian and Cyclist categories have an IoU threshold of 0.5.


\noindent \textbf{Waymo Open Dataset.}
The Waymo Open Dataset \cite{sun2020scalability} consists of 798 training sequences and 202 validation sequences for vehicles and pedestrians.
We experiment on this dataset to showcase the generalization of our method.
Due to the large dataset size and high frame rate \cite{reading2021categorical}, we use only a subset of the training and validation sets to form our training (3,000 samples) and validation (2,716 samples) sets, with only front camera images.
We scale images to size $960 \times 640$, and make labels along the KITTI standard, under which there are no `hard' category objects.
For evaluation, we use 0.7 IoU threshold for the Car category.

\noindent \textbf{Implementation.}
The network is trained on a single Nvidia Tesla V100S GPU.
We adopt an Adam optimizer with an initial learning rate of 2.25e-4 with a one-cycle learning rate policy, which gradually increases the learning rate to 2.25e-3. 
We train our model for 200 epochs with a batch size of 8.
To maintain the real-time performance, we adopt the $\frac{1}{16}$ size features to generate 3D coordinates grid.
The depth grid ranges from 2m to 46.8m for KITTI dataset and 2m to 55.76m for Waymo dataset, both with 64 depth bins.

\subsection{Main Results}
\noindent \textbf{Results in Car category on the KITTI test set.}
Table \ref{table_kitti_car_compare} shows the results of our method on the KITTI test set.
It compares our method with the state-of-the-art published monocular methods, where the ranking is based on the $AP_{3D}$ metrics at the moderate level. 
Our proposed method obtains the first rank place among these methods. 
Compared to the 2nd ranked one, we achieves an improvement of [+1.03\%, +0.71\%, +0.19\%] on $AP_{3D}$ and [+0.42\%, +0.55\%, +0.58\%] on $AP_{BEV}$. 
In addition, our method outperforms the baseline with large margins of [+2.18\%, +0.71\%, +0.19\%] on $AP_{3D}$ and [+1.38\%, +1.04\%, +1.30\%] on $AP_{BEV}$, which proves the effectiveness of our network design.

\begin{table}[t]
\centering
\caption{Comparison on the KITTI  test set. Highest result is marked with \textcolor{red}{red} and the second highest is marked with \textcolor{blue}{blue}. IoU = 0.7 is adopted as the threshold for the 'Car' category. `Improvement' indicates the offset with MonoCon \cite{liu2021learning}.
}
\vspace{-0.1in}
\setlength{\tabcolsep}{0.7mm}{
    \begin{tabular}{l|c|ccc|ccc}
        \toprule
        \multirow{2}{*}{Method} & \multirow{2}{*}{Times(ms)} & \multicolumn{3}{c|}{Car $AP_{3D}$}  & \multicolumn{3}{c}{Car $AP_{BEV}$}  \\
        &  & Easy & \textbf{Mod.} & Hard & Easy & \textbf{Mod.} & Hard\\ 
        \midrule
        SMOKE \cite{liu2020smoke}   & 30 & 14.03 & 9.76 & 7.84 &20.83 &14.49 &12.75\\
        PatchNet \cite{ma2020rethinking}  & 400 & 15.68 & 11.12 & 10.17 &22.97 &16.86 &14.97\\
        PGD \cite{wang2021probabilistic}  & 21 &19.05 &11.76 &9.39 &26.89 &16.51 &13.49\\
        MonoDLE \cite{ma2021delving}    & 40 & 17.23  & 12.26  & 10.29 & 24.79  & 18.89  & 16.00\\
        PCT \cite{wang2021progressive}  & 45 &21.00 &13.37 &11.31 &29.65 &19.03 &15.92 \\
        CaDDN \cite{reading2021categorical}   & 630 & 19.17 & 13.41 & 11.46 &27.94 &18.91 &17.19\\ 
        MonoEF \cite{zhou2021monocular}  & 30 & 21.29 & 13.87 & 11.71 &29.03 &19.70 &17.26\\ 
        MonoFlex \cite{zhang2021objects}   & 35 & 19.94 & 13.89 & 12.07 &28.23 &19.75 &16.89\\  
        AutoShape \cite{liu2021autoshape}  & 50 & 22.47 & 14.17 & 11.36 &30.66 &20.08 &15.95\\ 
        GUPNet \cite{lu2021geometry}   & 34 & 20.11 & 14.20 & 11.77  &- &- &-\\
        Homo \cite{gu2022homography} & - & 21.75  & 14.94  & 13.07& 29.60  & 20.68  & 17.81\\ 
        MonoDTR \cite{huang2022monodtr}  & 37 & 21.99 & 15.39 & 12.73 & 28.59 & 20.38 & 17.14\\
        MonoDETR \cite{zhang2022monodetr}   & - & \textcolor{blue}{23.65} & 15.92 & 12.99 & \textcolor{blue}{32.08} &  21.44  & 17.85 \\ 
        MonoDistill \cite{chong2022monodistill}  & 40 & 22.97 & 16.03 & 13.60 & 31.87 & \textcolor{blue}{22.59} & \textcolor{blue}{19.72} \\ 
        MonoCon \cite{liu2021learning}  & 26 & 22.50  & \textcolor{blue}{16.46}  & \textcolor{blue}{13.95}& 31.12 & 22.10 & 19.00 \\
        \midrule
        \textbf{Ours}  & 46   
        & \textbf{\textcolor{red}{24.68}} & \textbf{\textcolor{red}{17.17}} & \textbf{\textcolor{red}{14.14}} &\textbf{\textcolor{red}{32.50}} &\textbf{\textcolor{red}{23.14}} &\textbf{\textcolor{red}{20.30}}\\ 
        Improvement  & - & +2.18 & +0.71 & +0.19 & +1.38 & +1.04 & +1.30\\
        \bottomrule
    \end{tabular}
}

\label{table_kitti_car_compare}
\vspace{-0.2in}
\end{table}

\noindent \textbf{Results in Pedestrian and Cyclist categories on the KITTI test set.}
We further present the performance of the Pedestrian and Cyclist categories, as shown in Table \ref{table_kitti_other_compare}. 
Our approach is superior to our baseline MonoCon \cite{liu2021learning} over 1\%, and shows competitive performance with other methods ranking within the top 3.

\noindent \textbf{Results in Car category on the subset of Waymo Open dataset.}
We conduct experiments on the subset of Waymo Open Dataset where we train 3,000 samples and test on the validation subset (2,716 samples), as shown in Table~\ref{table_all_compare_waymo}.
We adopt the official codes of CaDDN \cite{reading2021categorical} and MonoCon \cite{liu2021learning} to perform experiments on the same subset of Waymo Open Dataset for a fair comparison.
From Table \ref{table_all_compare_waymo}, our method outperforms MonoCon \cite{liu2021learning} with margins of [+3.18\%, +2.90\%] on Car $AP_{3D}$ and [+1.01\%, +2.72\%] on Car $AP_{BEV}$, which indicates great generalization.

\begin{table}[t]
\centering
 \caption{Comparison on the KITTI  test set. Highest result is marked with \textcolor{red}{red} and the second highest is marked with \textcolor{blue}{blue}.
 IoU = 0.5 is used for 'Cyclist' and 'Pedestrian' categories. 
 `Improvement' indicates the offset with MonoCon \cite{liu2021learning}.
}
\vspace{-0.1in}
\setlength{\tabcolsep}{0.7mm}{
\begin{tabular}{l|ccc|ccc}
  \toprule
  \multirow{2}{*}{Method}  &  \multicolumn{3}{c|}{Pedestrian $AP_{3D}$}  &  \multicolumn{3}{c}{Cyclist $AP_{3D}$}  \\
   & Easy & Mod. & Hard & Easy & Mod. & Hard\\ 
   \midrule
   
   MonoDLE \cite{ma2021delving} & 9.64 & 6.55  & 5.44 
                                & 4.59  & 2.66  & 2.45\\
   CaDDN \cite{reading2021categorical}    & 12.87 & 8.14 & 6.76 & \textcolor{red}{7.00} & \textcolor{blue}{3.41} & \textcolor{red}{3.30} \\ 
   MonoEF \cite{zhou2021monocular}  & 4.27 & 2.79 & 2.21 & 1.80 & 0.92 & 0.71 \\ 
   MonoFlex \cite{zhang2021objects}  & 9.43 & 6.31 & 5.26 & 4.17 & 2.35 & 2.04 \\  
   GUP Net \cite{lu2021geometry}  & \textcolor{blue}{14.72} & 9.53 & 7.87 & 4.18 & 2.65 & 2.09 \\

    Homo \cite{gu2022homography}& 11.87  & 7.66  & 6.82
                                & 5.48  & \textcolor{red}{3.50}  & 2.99 \\ 
    MonoDTR \cite{huang2022monodtr}   & \textcolor{red}{15.33} & \textcolor{red}{10.18} & \textcolor{red}{8.61} 
                                    & 5.05 & 3.27 & \textcolor{blue}{3.19} \\
    MonoDistill \cite{chong2022monodistill}  & 12.79  & 8.17 & 7.45
                                            & 5.53 & 2.81 &  2.40 \\ 
    MonoCon \cite{liu2021learning}  & 13.10  & 8.41  & 6.94 
                                    & 2.80  & 1.92  & 1.55 \\ 
   \midrule
     \textbf{Ours}    
      & \textbf{14.16} & \textbf{\textcolor{blue}{9.67}} & \textbf{\textcolor{blue}{8.26}} & \textbf{\textcolor{blue}{5.88}} & \textbf{3.30} & \textbf{2.85} \\ 
     Improvement   & +1.06 & +1.26 & +1.32 & +3.08 & +1.38 & +1.30 \\

  \bottomrule
 \end{tabular}
 }

 \label{table_kitti_other_compare}
\end{table}

\begin{table}[t]
\centering
\caption{Comparison on the subset of Waymo Dataset. Highest result is marked with \textcolor{red}{red} and the second highest is marked with \textcolor{blue}{blue}. We adopt IoU =0.7 as the threshold for 'Car' category.
}
\vspace{-0.1in}
\setlength{\tabcolsep}{0.7mm}{
\begin{tabular}{l|cc|cc}
  \toprule
  \multirow{2}{*}{Method} &  \multicolumn{2}{c|}{Car $AP_{3D}$} &  \multicolumn{2}{c}{Car $AP_{BEV}$}  \\
  & Easy & Mod.   & Easy & Mod. \\ 
  \midrule

  CaDDN \cite{reading2021categorical}   &61.48    &54.33    &67.29 &57.09 \\ 

  MonoCon \cite{liu2021learning}  &\textcolor{blue}{64.24}  &\textcolor{blue}{56.17}      &\textcolor{blue}{69.77} &\textcolor{blue}{59.64}  \\ 
  \midrule
     \textbf{Ours}    
     & \textbf{\textcolor{red}{67.42}} & \textbf{\textcolor{red}{59.07}}   &\textbf{\textcolor{red}{70.78}} &\textbf{\textcolor{red}{62.36}} \\ 
     Improvement  & +3.18 & +2.90   & +1.01 & +2.72 \\
  \bottomrule

 \end{tabular}
 }
 \vspace{-0.2in}
 \label{table_all_compare_waymo}
\end{table}

\begin{figure*}[!t]
    \centering
    \includegraphics[width=0.8\textwidth]{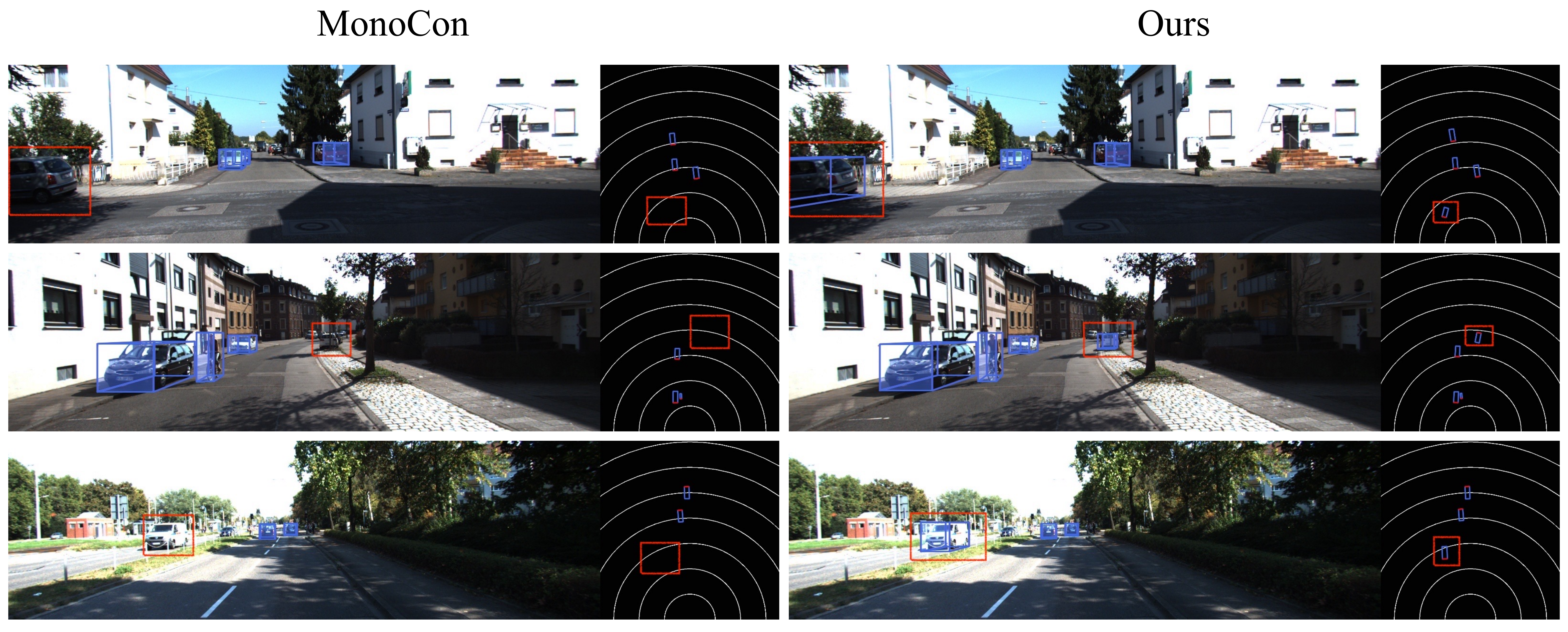}
    \vspace{-0.1in}
    \caption{Qualitative results on KITTI val set. 
    The blue bounding box denotes the predicted results.
    In the BEV image, the red side denotes the cars' head.
    Our proposed method dominates the baseline.}
    \label{fig:quality2}
    \vspace{-0.2in}
\end{figure*}

\noindent \textbf{Time analysis.}
We measure our model's average inference time on a single Nvidia Tesla V100S GPU.  
As shown in Table \ref{table_kitti_car_compare}, our model can achieve real-time performance at 25 FPS, which validates the efficiency of our framework. 
Specifically, our MonoPGC runs 15.75$\times$ faster than CaDDN \cite{reading2021categorical}, for our method avoids time-consuming lifting module.
Our pixel depth estimation and transformer bring some additional computation with a little slower speed than MonoCon \cite{liu2021learning}, but our result greatly overcomes the MonoCon \cite{liu2021learning}.

\subsection{Ablation Study}
We conduct ablation studies on the KITTI validation split with models trained on the training split.

\noindent \textbf{Importance of each proposed components.} 
Table \ref{ablation_total} shows the ablation studies about our method, and our three modules perform improving the detection performance.
When adding depth cross-attention pyramid module (c) and depth-gradient positional encoding (d), our $AP_{3D}$ and $AP_{BEV}$ both increase especially in the easy category, since DCPM enhances more robust depth-aware features and DCPE strengthens the unitary depth information, which enriches the pixel geometry contexts for 3D detection.
Setting (b) introduces an individual depth-space-aware transformer where 2D visual features display the transformer's input.
Without depth knowledge and other surround-view images to make up the scene information, 2D features can't adapt to the 3D position well, making little influence on our baseline.
However, after introducing DCRM (e), depth-aware features match well with the space coordinates.   
Finally, our method achieves the highest accuracy applying all modules.

\begin{table}[t]
\centering
\caption{Ablation studies of each proposed components.}
\vspace{-0.1in}
\setlength{\tabcolsep}{0.9mm}{
\begin{tabular}{c|ccc|ccc|ccc}
 \Xhline{1pt}
 & \multirow{2}{*}{DCPM}
 & \multirow{2}{*}{DCPE} 
     & \multirow{2}{*}{DSAT} 
     & \multicolumn{3}{c|}{$AP_{3D}$@IoU=0.7} 
     & \multicolumn{3}{c}{$AP_{BEV}$@IoU=0.7}\\  \cline{5-10}
     & & & & Easy & Mod. & Hard & Easy & Mod. & Hard \\ \hline
    (a) & - & -  & -  &22.88 &17.81 &15.19 &31.17 &23.78 &20.58\\
    (b) & - & - & \checkmark & 22.85 & 17.84 & 15.12 & 30.96 & 23.57 & 20.57 \\ 
    (c) & \checkmark & - & - & 24.73 & 17.98 & 15.25& 33.26 & 23.91 & 20.61 \\
    (d) &\checkmark &\checkmark & - & 24.91 & 18.18 & 15.31 & 33.28 & 23.96 & 20.60 \\
    (e) &\checkmark & -  &\checkmark  & 25.42 & 18.62 & 15.60 & 33.40 & 24.10 & 20.58\\
    (f) &\checkmark & \checkmark &\checkmark & \textbf{25.67} & \textbf{18.63} & \textbf{15.65} & \textbf{34.06} & \textbf{24.26} & \textbf{20.78}\\
    \hline
 \Xhline{1pt}
\end{tabular}
}

\label{ablation_total}
\end{table}
\begin{table}[t]
\centering
\caption{
Ablation studies of different coordinates grid setting on the KITTI validation set. `d.v.' denotes the direction vector.
}
\vspace{-0.1in}
\setlength{\tabcolsep}{0.75mm}{
\begin{tabular}{l|ccc|ccc}
 \Xhline{1pt}
    \multirow{2}{*}{Ablation} & \multicolumn{3}{c|}{$AP_{3D}$@IoU=0.7} 
     & \multicolumn{3}{c}{$AP_{BEV}$@IoU=0.7}\\  \cline{2-7}
      & Easy & Mod. & Hard & Easy & Mod. & Hard \\ \hline
    (a) No coord. grid  &22.88 &17.81 &15.19 &31.17 &23.78 &20.58 \\
    (b) d.v. \cite{zhou2022cross} w/o DCPF &22.85 &17.87 &15.22 &31.24 &23.81 &20.60\\
    (c) d.v. \cite{zhou2022cross} w DCPF &24.77 &18.16 &15.28 &33.00 &23.85 &20.62\\
    (d) 3D coord. grid w/o DCPF  &22.85 &17.84 &15.12 &30.96 &23.57 &20.57\\
    (e) 3D coord. grid w DCPF  & \textbf{25.67} & \textbf{18.63} & \textbf{15.65} & \textbf{34.06} & \textbf{24.26} & \textbf{20.78}\\

    \hline
 \Xhline{1pt}
 
\end{tabular}
}
\label{ablation_3d_grid}
\vspace{-0.2in}
\end{table}

\begin{table}[t]
\centering
\caption{
Ablation studies of different positional encodings on the KITTI validation set.
}
\vspace{-0.1in}
\setlength{\tabcolsep}{1.5mm}{
\begin{tabular}{l|ccc|ccc}
 \Xhline{1pt}
  
    \multirow{2}{*}{Method} & \multicolumn{3}{c|}{$AP_{3D}$@IoU=0.7} 
     & \multicolumn{3}{c}{$AP_{BEV}$@IoU=0.7}\\  \cline{2-7}
      & Easy & Mod. & Hard & Easy & Mod. & Hard \\ \hline
    No PE   &23.23 &17.33 &14.74 &31.32 &23.26 &20.07\\
    Sinusoidal \cite{vaswani2017attention}  & 24.78 & 18.17 & 15.09 & 33.00 & 23.85 & 20.33\\
    APE \cite{VIT}  & 24.34 & 18.36 & 15.31 & 32.42 &23.98 &  20.41 \\
    DPE \cite{huang2022monodtr}  & 25.42 & 18.50 &  15.57 & 33.48 & 24.00 & 20.59 \\
    DGPE   & \textbf{25.67} & \textbf{18.63} & \textbf{15.65} & \textbf{34.06} & \textbf{24.26} & \textbf{20.78}\\

    \hline
 \Xhline{1pt}
\end{tabular}
}

\label{ablation_PE}
\vspace{-0.1in}
\end{table}

\noindent \textbf{Comparison with different coordinates grid setting.}
We investigate different coordinates grid setting with/without depth cross-attention pyramid features (DCPF) in monocular detection, as shown in Table \ref{ablation_3d_grid}.
We refer to the work \cite{zhou2022cross} to introduce the direction vector as the transformer encoder's input.
Setting (a) denotes our baseline.
From Table~\ref{ablation_3d_grid}, depth-aware DCPF could maximize the potential of spatial encoding (c)(e), 
but the direction vector method doesn't perform superior to our 3D coordinates grid. 
Its depth 1 setting may limit the feature enhancement and the monocular detection task doesn't have other surround-view images to assist scene understanding. 
Setting (b) and (d) also reveal that visual features perform not well with spatial encoding in monocular detection task.

\noindent \textbf{Comparison with different positional encodings.}
We discuss the effects of the proposed depth-gradient positional
encoding (DGPE) in Table \ref{ablation_PE}.
We compare our DGPE with several previous positional encodings, including without
using positional encoding (No PE), sinusoidal positional encoding \cite{vaswani2017attention}, absolute positional encoding (APE) \cite{VIT}, and depth positional encoding (DPE) \cite{huang2022monodtr}. 
Our proposed DGPE achieves better results on KITTI validation set, for DGPE extracts more meticulous and attentive clues with the gradient-based filter, which enhances the pixel-level geometry information compared to the other encodings.

\subsection{Qualitative Results}
We provide the qualitative examples on the KITTI validation set, as shown in Figure \ref{fig:quality2}.
Compared with MonoCon \cite{liu2021learning}, our method's prediction reveals more accurate and robust.
MonoCon \cite{liu2021learning}, short of geometry clues, fails to detect vehicles affected by too dark or too bright illumination and occlusion.
Our method succeeds in these hard cases, contributed from the features' sufficient pixel geometry contexts.
The supplementary material includes more qualitative results.

\section{Conclusion}
In this paper, we propose \textbf{MonoPGC}, a novel end-to-end monocular 3D object detection framework with adequate pixel geometry contexts.
The proposed depth cross-attention pyramid module (\textbf{DCPM}) expands the receptive field enabling a more comprehensive and precise perception of depth geometry content.
The depth-space-aware transformer (\textbf{DSAT}) efficiently incorporate 3D geometry position with DCPF, mapping geometry contents to pixel features.
Finally depth-gradient positional encoding (\textbf{DGPE}) acts as a valuable indication for anchor-free object detection.
Extensive experiments on the KITTI dataset validate that our model achieves the state-of-the-art performance on the KITTI dataset and a portion of the Waymo Open dataset.
In future work, we would like to explore novel monocular architecture without the explicit depth estimation. 

\section*{APPENDIX}
\subsection{The details of our depth cross-attention pyramid module (DCPM)}
\begin{figure}[ht]
    \centering
    \includegraphics[width=0.48\textwidth]{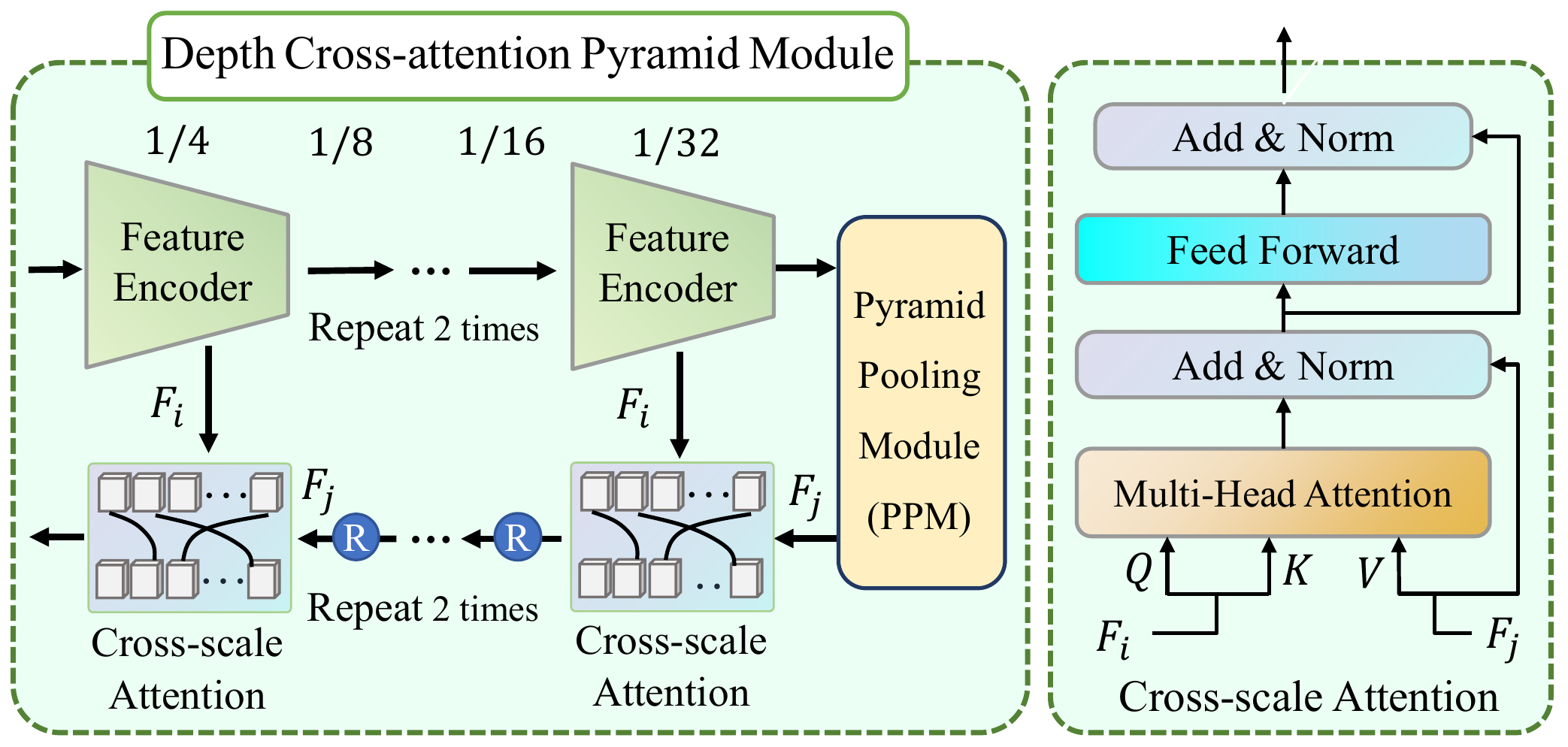}
    \caption{The details of our Depth Cross-attention Pyramid Module (DCPM), which adequately integrates multi-level image features via feature encoders, cross-scale attention layers, and PPM module. 
    ``$R$" denotes the rearranged up-scaling for feature maps.}
    \label{fig:dcpm}
\end{figure}

As shown in Fig.~\ref{fig:dcpm}, we deliver more details of the depth cross-attention pyramid module (DCPM).
Specifically, we adopt the backbone to extract four different-scale features.
The low-level features focus on the local details, while the high-level features seize the global context information.
Afterward, we employ the pyramid pooling module (PPM) to aggregate these features and deliver four cross-scale attention layers to fuse the multi-level feature maps.
We use the scaled features $F_i$ as the query and key, and adopt the fusion features $F_j$ as the value to stimulate the interaction of multi-scale features. Within each cross-attention layer, we also introduce the rearranged up-scaling operation \cite{yuan2022newcrfs} to reduce the network complexity and boundary contour refinements.

\subsection{Transformation from camera frustum space to 3D world space.}
\begin{figure}[ht]
    \centering
    \includegraphics[width=0.43\textwidth]{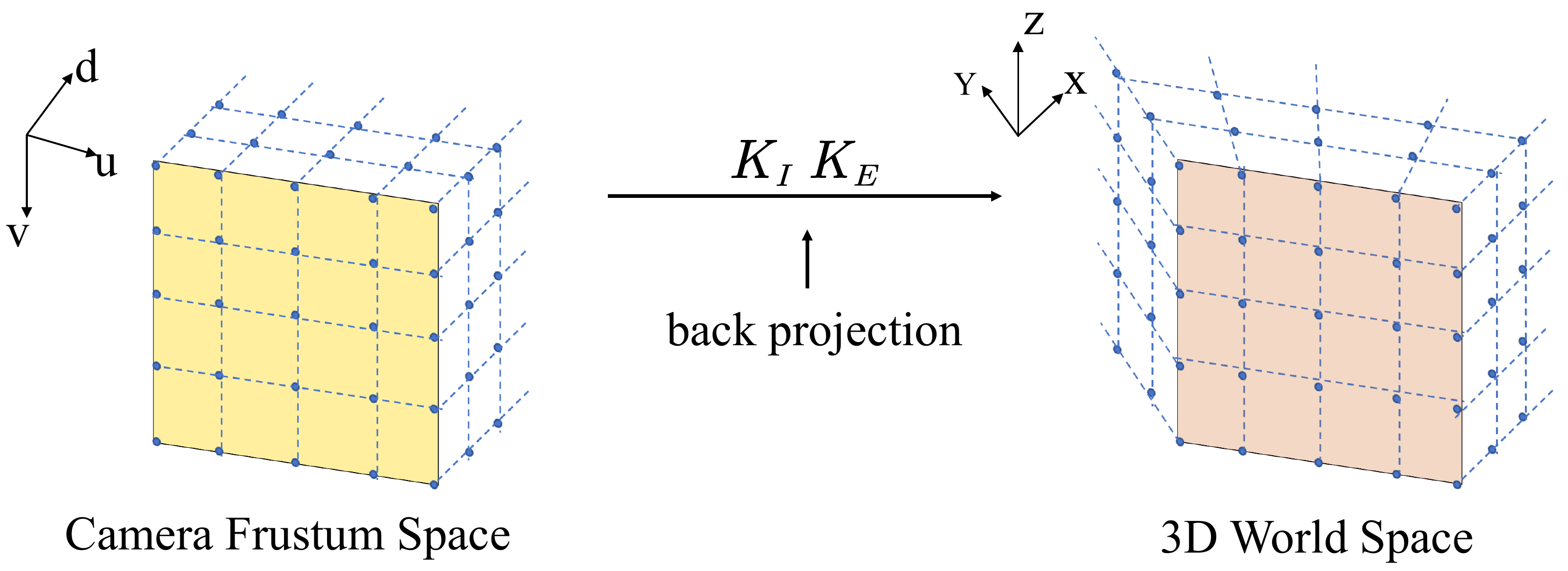}
    \caption{The back projection from camera frustum space to 3D world space. $K_{E}$ and $K_{I}$ indicate the extrinsics and intrinsics calibration matrix.}
    \label{fig:generator}
\end{figure}

We need to reversely project our camera frustum space points to the 3D world space with the intrinsic and extrinsic calibration \cite{reading2021categorical,zhou2022cross}:
\begin{equation}
P_{j}^{3d}=K_{E}^{-1}K_{I}^{-1}P_{j}^{m}
\end{equation}
where $P_{j}^{m}\in R^{4}$ denotes the camera frustum space coordinate and $P_{j}^{3d}\in R^{4}$ denotes the 3D world space coordinate. $K_{E}\in R^{4\times 4}$ and $K_{I}\in R^{4\times 4}$ indicate the extrinsics and intrinsics calibration matrix, and the transformation process demonstrates in Figure \ref{fig:generator}.

\subsection{Comparison of different settings with depth cross-attention pyramid module (DCPM)}
We investigate different settings about the multi-scale features and cross-attention layers in DCPM on the KITTI \cite{kitti} validation set, as shown in Table \ref{ablation_DCPM}.
Setting (a) denotes that we adopt the single-scale feature ($\frac{1}{4}$ scale feature) and MLPs layer to produce depth-aware features, which performs an inferior result.
Multi-scale features with the multi-scale receptive field, help the detection task improve, from setting (b). 
Setting (c) shows that cross-attention is more suitable for the multi-level features fusion with more efficient cooperation with global and local features, compared with MLPs layer. 
Finally, our method achieves the highest accuracy by applying two settings.

\begin{table}[h]
\centering
\caption{Comparison of different settings with depth cross-attention pyramid module (DCPM) on the KITTI \cite{kitti} validation set. `MS' denotes `Multi-scale' and `CA' denotes `Cross-attention'.}
\setlength{\tabcolsep}{1.8mm}{
\begin{tabular}{c|cc|ccc|ccc}
 \Xhline{1pt}
 & \multirow{2}{*}{MS}
 & \multirow{2}{*}{CA} 
     & \multicolumn{3}{c|}{$AP_{3D}$@IoU=0.7} 
     & \multicolumn{3}{c}{$AP_{BEV}$@IoU=0.7}\\  \cline{4-9}
     &  & & Easy & Mod. & Hard & Easy & Mod. & Hard \\ \hline
    (a) & -   & -  &23.29 &18.23 &15.39 &31.68 &23.64 &20.60\\
    (b) & -  & \checkmark & 24.66 & 18.32 & 15.46 & 32.50 & 23.89 & 20.65 \\ 
    (c) & \checkmark & -  & 24.83 & 18.46 & 15.55& 33.34 & 24.08 & 20.71 \\
    (d) &\checkmark & \checkmark & \textbf{25.67} & \textbf{18.63} & \textbf{15.65} & \textbf{34.06} & \textbf{24.26} & \textbf{20.78}\\
    \hline
 \Xhline{1pt}
\end{tabular}
}
\label{ablation_DCPM}
\end{table}

\subsection{More comparative qualitative results.}

\begin{figure*}[!t]
    \centering
    \includegraphics[width=1\textwidth,height=0.8\textwidth]{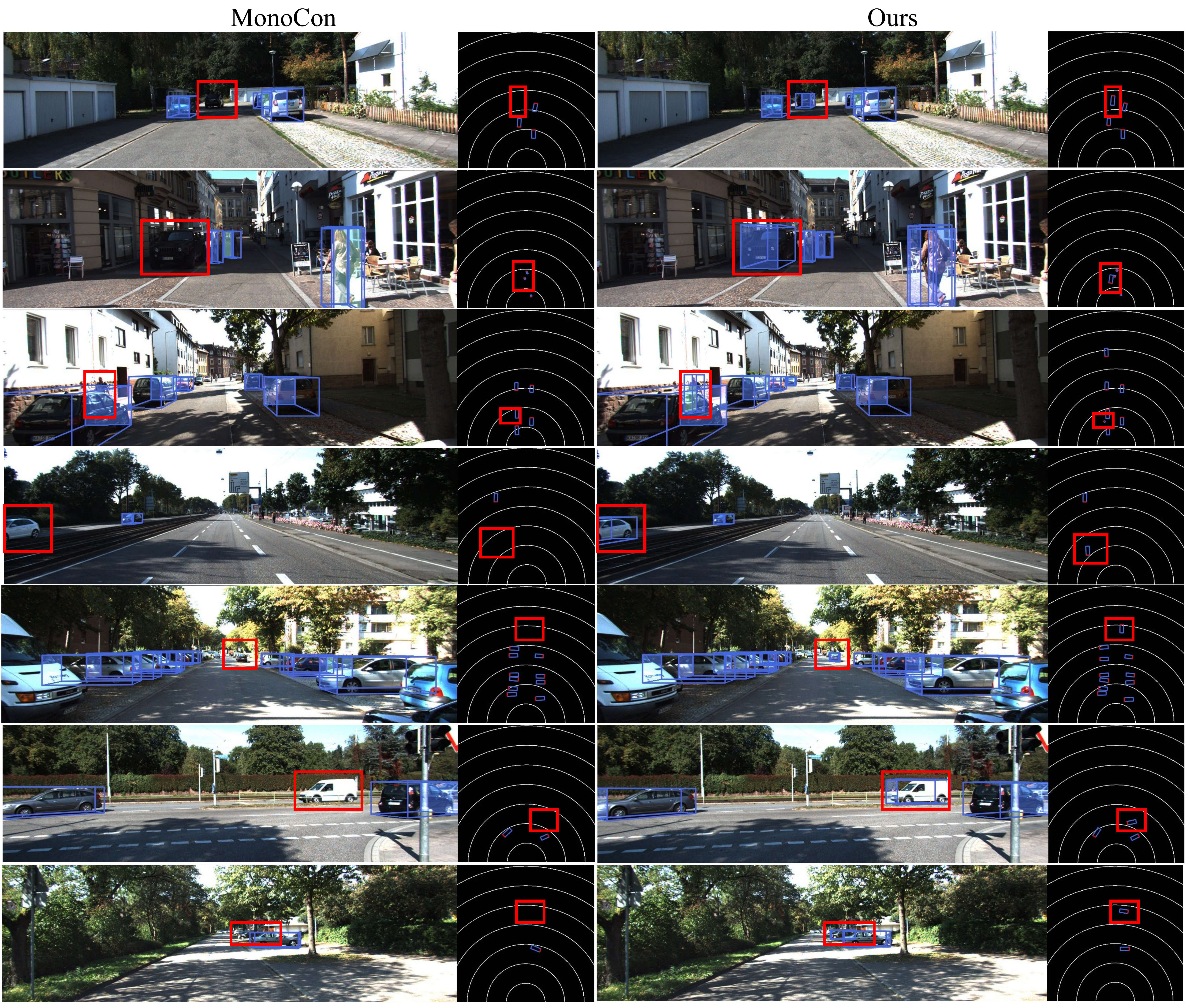}
    \caption{More qualitative results on KITTI \cite{kitti} validation set. 
    The blue bounding box denotes the predicted results.
    In the BEV image, the red side denotes the cars' head.
    Our proposed method dominates the baseline with too dark or too bright illumination, truncated, occlusion and too far away objects occasions.}
    \label{fig:quality_all}
\end{figure*}

We provide more qualitative examples on the KITTI \cite{kitti} validation set, as shown in Figure \ref{fig:quality_all}.
Compared with MonoCon \cite{liu2021learning}, our method better tackles the tough detection occasions, like too dark or too bright illumination, truncated, occlusion and too far away objects.
Our method succeeds in these hard cases, contributing to the features' sufficient pixel geometry contexts.

\subsection{Additional Details.}
We benefit from the awesome works of MonoCon \cite{liu2021learning}, MonoDLE \cite{ma2021delving} and CaDDN \cite{reading2021categorical}.
Our data augmentation and training config refers to MonoCon \cite{liu2021learning}.
The depth bin setting refers to CaDDN \cite{reading2021categorical}.

\bibliographystyle{IEEEtran}

\end{document}